\documentclass{article}
\usepackage{enumitem}
\usepackage{siunitx}
\usepackage{booktabs}
\usepackage{tablefootnote}
\usepackage[hang,flushmargin]{footmisc}
\usepackage{amssymb}

\usepackage{spconf,tabulary,graphicx,amsmath,amsfonts,amssymb,setspace}
\usepackage[figurename=Fig.]{caption}
\usepackage{booktabs}
\usepackage{tabularx}
\usepackage{multirow}
\usepackage{float}
\usepackage{subfig}
\usepackage{bm}

\usepackage{colortbl} 
\usepackage{xcolor}
\usepackage{pifont}

\usepackage[pagebackref=false,breaklinks=true,letterpaper=true,colorlinks=true,bookmarks=false,urlcolor=black, 
citecolor=blue
]{hyperref}
\newlength\savewidth\newcommand\shline{\noalign{\global\savewidth\arrayrulewidth
  \global\arrayrulewidth 1pt}\hline\noalign{\global\arrayrulewidth\savewidth}}

\newlength\thinwidth

\definecolor{Gray}{gray}{0.92}
\definecolor{DarkGray}{gray}{0.5}
\newcolumntype{x}{>{\columncolor{Gray}}c}
\newcolumntype{H}{>{\setbox0=\hbox\bgroup}c<{\egroup}@{}}
\definecolor{LightCyan}{rgb}{0.88,1,1}
\definecolor{altRowColor}{gray}{0.92}
\definecolor{highlightRowColor}{rgb}{0.9, 0.9, 1}

\definecolor{GrayNumber}{gray}{0.5}
\definecolor{GrayXMark}{gray}{0.7}
\newcommand{\cmark}{\ding{51}}%
\newcommand{\xmark}{ {\color{GrayXMark} \ding{55}} } %

\definecolor{ImageDark}{rgb}{0,0.3,0.8}
\definecolor{VideoDark}{rgb}{.5,.0,.5}
\definecolor{DepthDark}{rgb}{0,.5,0}
\definecolor{AudioDark}{rgb}{0.11764705882352941, 0.5647058823529412, 1.0}
\definecolor{ThermalDark}{rgb}{0.8823529411,0.63725490196,0.0156862745}
\definecolor{IMUDark}{rgb}{0.6235294117647059, 0.27058823529411763, 0.4627450980392157}
\colorlet{Image}{ImageDark!20!white}
\colorlet{Video}{VideoDark!20!white}
\colorlet{Depth}{DepthDark!20!white}
\colorlet{Audio}{AudioDark!20!white}
\colorlet{ImageLight}{ImageDark!70!white}
\colorlet{VideoLight}{VideoDark!70!white}
\colorlet{DepthLight}{DepthDark!70!white}
\colorlet{AudioLight}{AudioDark!70!white}

\newcolumntype{i}{>{\columncolor{Image}}c}
\newcolumntype{v}{>{\columncolor{Video}}c}
\newcolumntype{d}{>{\columncolor{Depth}}c}
\newcolumntype{a}{>{\columncolor{Audio}}c}
\newcolumntype{I}{>{\columncolor{ImageLight}}c}
\newcolumntype{V}{>{\columncolor{VideoLight}}c}
\newcolumntype{D}{>{\columncolor{DepthLight}}c}
\newcolumntype{A}{>{\columncolor{AudioLight}}c}
\newcolumntype{E}{>{\columncolor{highlightRowColor}}c}

\title{Bridging the Gap between Text, Audio, Image, and Any Sequence: \\A Novel Approach using Gloss-based Annotation}
\name{\begin{tabular}{c}Sen Fang$^{1}$, Sizhou Chen$^{2, \star}$, Yalin Feng$^{3}$, Xiaofeng Zhang$^{4}$, Teik Toe Teoh$^{5}$\end{tabular}
\thanks{$^{\star}$ Collaborator Author. This article is a technical report and was not intended to be published due to my waning interest in the field.
}
}
\address{$^{1,3}$Victoria University, $^{2}$Chengdu University of Information Technology, \\$^{4}$Shanghai Jiao Tong University, $^{5}$Nanyang Technological University
}

\begin{document}
\ninept
\maketitle
\begin{abstract}
This paper presents an innovative approach called \textbf{BGTAI} to simplify multimodal understanding by utilizing gloss-based annotation as an intermediate step in aligning \textbf{T}ext and \textbf{A}udio with \textbf{I}mages. While the dynamic temporal factors in textual and audio inputs contain various predicate adjectives that influence the meaning of the entire sentence, images, on the other hand, present static scenes. By representing text and audio as gloss notations that omit complex semantic nuances, a better alignment with images can potentially be achieved. This study explores the feasibility of this idea, specifically, we first propose the first \textbf{Langue2Gloss} model and then integrate it into the multimodal model UniBriVL for joint training. To strengthen the adaptability of gloss with text/audio and overcome the efficiency and instability issues in multimodal training, we propose a \textbf{DS-Net} (Data-Pair Selection Network), an \textbf{Result Filter} module, and a novel \textbf{SP-Loss} function. Our approach outperforms previous multimodal models in the main experiments, demonstrating its efficacy in enhancing multimodal representations and improving compatibility among text, audio, visual, and any sequence modalities.
\end{abstract}



\section{Introduction}
\vspace{-4pt}
In recent years, there has been a growing trend towards the integration of multiple modalities such as language, audio, and vision. Numerous studies \cite{fei2022towards,wang2022image,girdhar2023imagebind,zhang2023metatransformer} have explored various methods to effectively fuse three or even up to twelve modalities into a unified representation. Capitalizing on the availability of large-scale datasets, these approaches primarily rely on massive pretraining to train models that can then be transferred to various downstream tasks, yielding impressive results. Most of these studies are based on transformer architectures or modified versions thereof. Depending on their specific usage and architectural details, they employ diverse fusion techniques to integrate the different modalities effectively.
The prevailing approach \cite{fei2022towards,wang2022image} involves jointly training based on specific architectures and then manually converting them into the desired task format. However, such parameters often fail to be effectively shared across modalities. The latest approach \cite{girdhar2023imagebind,zhang2023metatransformer} aims to achieve parameter sharing and structural generalization by employing a universalized architecture, such as sharing encoders or utilizing multi-path transformers. This facilitates task transferability and enhances the usability of the shared structure. Nevertheless, previous methods have not effectively achieved parameter sharing among multiple modalities, as they only store them within a single model.
\begin{figure}[ht]
\includegraphics[width=\linewidth]{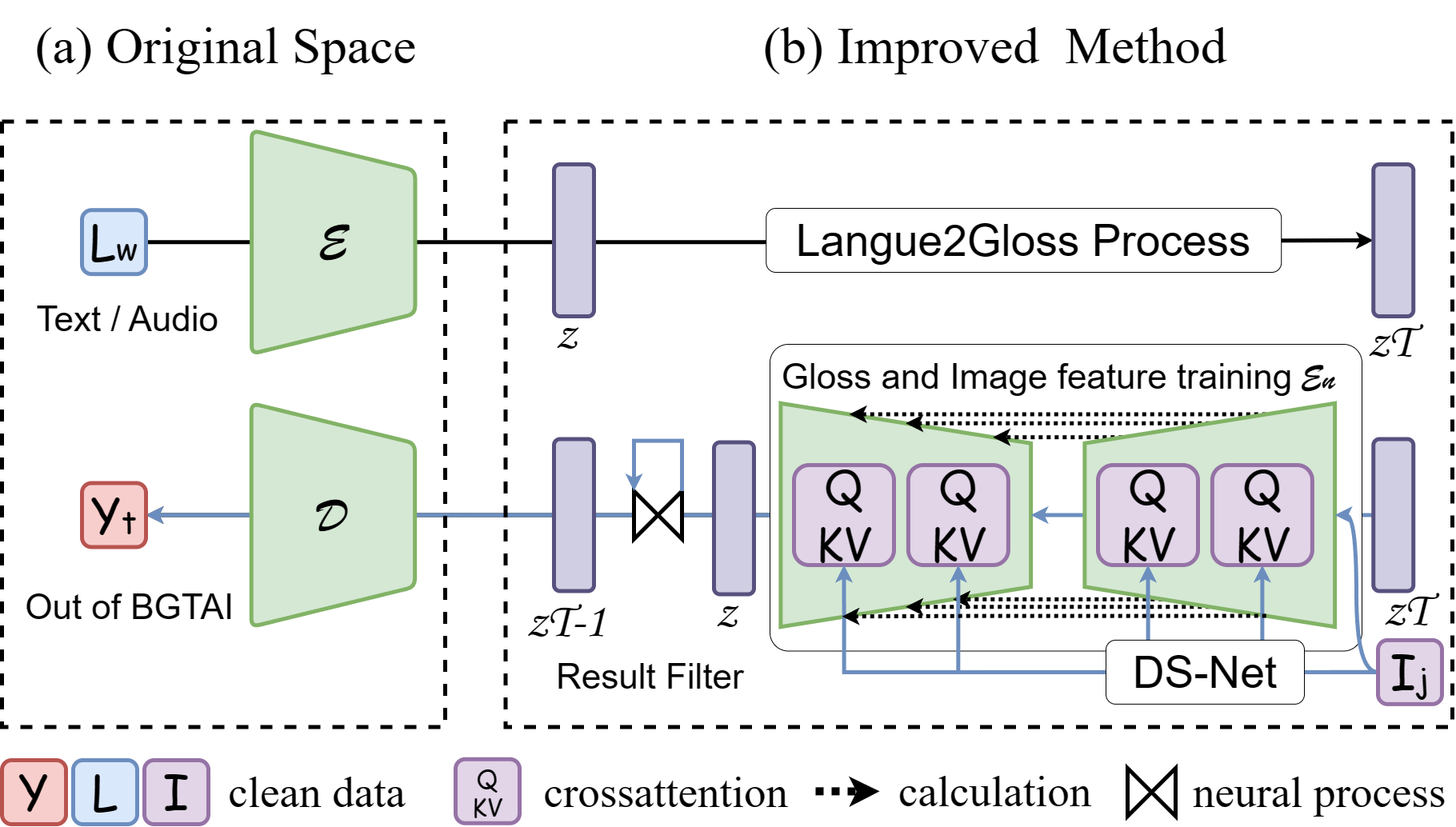}
\caption{(a) BGTAI Production Pipeline and ((b) Structure Drawing. It reveals a structure that unifies all representation possibilities.}
\label{fig:methods}
\vspace{-16pt}
\end{figure}

Gloss is an annotation method \cite{f2e24cc3-4208-3558-92fd-a3aa430647aa} applied in sign language, where certain postures or gestures (essentially sequences) can be represented as Gloss. In other words, it serves as an intermediary medium connecting natural language and complex sequences. In the field of sign language, such as sign language translation or sign language generation, utilize Gloss to enhance their model capabilities \cite{camgoz2020sign,Saunders_2022_CVPR}. Moreover, in reality, the application of Gloss has been employed for several decades to enhance language-gesture learning abilities for individuals with hearing impairments \cite{f2e24cc3-4208-3558-92fd-a3aa430647aa}. Therefore, we believe that incorporating real-world scenarios into multimodal learning can yield promising outcomes, based on the highly beneficial and validated methods.
UniBriVL is a state-of-the-art multimodal model \cite{fang2023unibrivl} for language-visual tasks that utilize a specialized encoder to jointly train text and audio, resulting in a universal language representation. This representation is then utilized to train the model for the task of matching with images. 
Our exploratory work focused on this model in the form of components.

In this study, we propose the \textbf{BGTAI} (Bridging the Gap between Text, Audio, Image, and any Sequence) framework aimed at addressing the disparities among text, audio, images, and any other sequences. The proposed framework comprises several components including the groundbreaking Langue2Gloss model, the DS-Net model, a Result Filter module, and a novel SP-Loss function, which are designed to solve some of the problems encountered in model training and application. We conducted a comprehensive evaluation involving various qualitative and quantitative assessments, such as baseline comparisons (including audio retrieval, audio classification, audio captioning, image retrieval, etc.), ablation experiments, and training efficiency tests. The experimental results show our method's superior effectiveness, surpassing previous similar approaches in multimodal tasks, and highlighting its significance for future multimodal model development and application.

\section{Proposed Method}

In this chapter, we provide a detailed introduction to our motivation and propose several components including Langue2Gloss, DS-Net, Result Filter, and SP-Loss as part of our method.

\vspace{-0.27cm}
\subsection{Motivation}
\vspace{-0.1cm}
The motivation behind our research is the need to address the limitations of existing multimodal training approaches in aligning image (A kind of sequence) with text/audio. To overcome these limitations, we propose the DS-Net, a Data-Pair Selection Network, which enhances alignment by selecting suitable training data pairs. We also introduce an improvement module to address efficiency and stability issues in multimodal training. Furthermore, we present a novel SP-Loss function that optimizes the training process and improves overall model performance. Our work also integrates the Langue2Gloss model into the UniBriVL for joint training, enhancing the adaptability of the image with langue. Through the incorporation of the DS-Net, Result Filter module, and SP-Loss function, we aim to provide more efficient and effective solutions for multimodal alignment.

\vspace{-0.27cm}
\subsection{Our components}
\vspace{-0.1cm}

\subsubsection{Langue2Gloss}
\vspace{-0.1cm}
Given an audio/text sentence, $L^N = \{w_1, w_2, ..., w_N\}$, use SpeechLM \cite{zhang2023speechlm} encoder (It can unify audio and text into a single representation) maps the sequence into a latent representation as follows:
\begin{equation}
o_{1:N}, h_L = Encoder(L^N)
\end{equation}
Here, $o_{1:N}$ represents the output of the encoder for each word w, and $h_L$ is the hidden representation of the provided sentence. The hidden representation and the encoder outputs are then passed to our decoder. The decoder utilizes an attention mechanism to generate sign gloss sequences, one gloss at a time:
\begin{equation}
gloss_m = Decoder(h_L, \alpha(o_{1:N}), gloss_{m-1})
\end{equation}

In the above equation, $\alpha (.)$ denotes the attention function, and $gloss_{m}$ is the gloss produced at time step m.
We ingeniously leverage the original text/audio inputs and the textually summarized/summarized audio outputs from the How2 dataset \cite{Sharma2022} as our training set, with Gloss serving as an intermediary.

\vspace{-0.27cm}
\subsubsection{DS-Net}
\vspace{-0.1cm}
To co-articulate between inputs and gloss, we propose a Data-Pair Selection Network (DS-Net) that learns to predict the temporal alignment between the original text/audio input and the summarized text/summary audio output. The DS-Net predicts a discrete sparse monotonic temporal alignment path, represented by a binary decision matrix $\hat{A}$ of size Q x T, where Q is the length of the summarized sequence and T is the length of the original sequence. 

This alignment path determines which data-pairs to select and which data-pairs to skip in order to generate a co-articulated continuous sequence. T is the length of the original sequence. 
Formally, the DS-Net predicts the alignment path as follows:
\begin{equation} \hat{A} = \text{DS-Net}(R, h_{1:\mathcal{W}}) \end{equation}
\begin{equation}\hat{Y} = Y \times \hat{A}\end{equation}

This operation allows us to map sequences of varying lengths, with the end of the sequence determined by the alignment selection of the final data-pair, which is built into the Langue2Gloss model.

\vspace{-0.34cm}
\subsubsection{Result Filter}
\vspace{-0.1cm}
DS-Net can optimize output for any sequence if glosses are replaced with images or any sequence by using the Dynamic Time Warping \cite{berndt1994dtw} (DTW) algorithm as a supervision signal, at this point we call this module the Result Filter module.
We pre-compute the DTW path, denoted as $\mathcal{A}^{*}$, between the interpolated text/audio dictionary sequence $\mathcal{I}$ and the target image continuous sequence $\mathcal{Y}$. 
To train the Result Filter, we compute a cross entropy loss $\mathcal{L}_{CE}$ between the predicted 1D temporal alignment, $\hat{\textrm{A}} \in \mathbb{R}^{\mathcal{Q}}$, and the ground truth DTW alignment, $\textrm{A}^{*} \in \mathbb{R}^{\mathcal{Q} \times 1}$, as:
\begin{equation}
    \mathcal{L}_{CE} (\hat{\textrm{A}},\textrm{A}^{*}) = - \frac{1}{\mathcal{Q}} \sum_{q=1}^{\mathcal{Q}} \textrm{A}^{*}_{q} \cdot{} \log (\hat{\textrm{A}}_{q})
\end{equation}

The final continuous image sequence, $\hat{\mathcal{Y}} = (y_{1},...,y_{\mathcal{T}})$, is produced as shown in Eq. \ref{fig:methods}. The first two components are also in it.

\vspace{-0.27cm}
\subsubsection{SP-Loss.}
\vspace{-0.1cm}
Specifically, the $\mathcal{A}^{*}$ and $\hat{\textrm{A}}$ obtained through the DTW algorithm can be utilized for calculating the SP-Loss.
To optimize the SP-Loss, we introduce a normalization step before computing the loss. Firstly, we normalize $\mathcal{A}^{*}$ and $\hat{\textrm{A}}$ to ensure that their values fall within the range of [0, 1]. Then, we use the normalized $\mathcal{A}^{*}$ and $\hat{\textrm{A}}$ to calculate the SP-Loss. The SP-Loss is computed as follows:
\begin{equation}
    \text{SP-Loss} = \frac{1}{\mathcal{Q}} \sum_{q=1}^{\mathcal{Q}} \left| \mathcal{A}^{*}_{q} - \hat{\textrm{A}}_{q} \right|
\end{equation}
Here, $\mathcal{Q}$ represents the length of the sequence when calculating.
The aforementioned process describes the computation of the optimized SP-Loss based on the input provided by the DTW algorithm. 

\section{Experiments}

In this section, we assess the performance and effectiveness of our model, including baseline comparison, training efficiency evaluation, ablation study, qualitative evaluation, and zero-shot comparison with previous works.

\vspace{-0.27cm}
\subsection{Experimental Setup}
\label{ssec:Setup}
\vspace{-0.1cm}

Table \ref{tab:audio_tasks} shows the datasets used in our primary downstream tasks and their corresponding details, encompassing classification tasks such as multi-class (MC), multi-label (ML), and zero-shot (ZS), retrieval tasks including audio retrieval (AR) and cross-modal retrieval (CMR), as well as the audio captioning (AC) task. 

\begin{table}[ht]
\centering
\resizebox{0.99\linewidth}{!}{
\begin{tabular}{@{}ccl@{}r@{}c@{}}\toprule
Dataset & Task & \# Clip (Split) & \# Class & Metric \\
\midrule
ESC-50 \cite{piczak2015dataset} & MC/ZS & 2k (5 folds) & 50 & ACC \\
UrbanSound8K \cite{Salamon:UrbanSound:ACMMM:14} & MC/ZS & 8k (10 folds) & 10 & ACC \\
FSD50K \cite{fonseca2020fsd50k} & ML/ZS & 50k & 200 & mAP \\
VGGSound \cite{chen2020vggsound} & MC/ZS & 185k & 309 & mAP \\
TAU Audio \cite{9415085} & MC/ZS & 190k & 10 & ACC \\
\midrule
DESED \cite{Turpault2019_DCASE} & AR & 2.5k (valid) & 10 & F1 \\
VGGSound \cite{chen2020vggsound} & CMR & 15k (test) & 309 & MRR \\
\midrule
Clotho \cite{Drossos_2020_icassp} & AC & 5k (evaluation) & & COCO\tablefootnote{https://github.com/tylin/coco-caption. Other datasets are mentioned later.} \cite{chen2015microsoft} \\
\bottomrule
\end{tabular}
}
\caption{Datasets for Downstream Tasks and Evolution Content.}
\label{tab:audio_tasks}
\vspace{-1.0em}
\end{table}

\begin{table*}[ht]
\centering
\begin{tabular}{@{}c@{\hskip 0.1in}c@{\hskip 0.1in}c@{\hskip 0.1in}c@{\hskip 0.1in}c@{\hskip 0.1in}c@{\hskip 0.1in}c@{\hskip 0.1in}c@{\hskip 0.1in}c@{}}\toprule
& \multicolumn{5}{c}{Classification} & \multicolumn{3}{c}{Retrieval} \\
\cmidrule(lr){2-6} \cmidrule(lr){7-9}
Model & ESC-50 & UrbanSound8K & FSD50K & VGGSound & TAU Audio Only & DESED (AR) & \multicolumn{2}{c}{VGGSound (CMR)} \\
\cmidrule(lr){8-9}
& ACC & ACC & mAP & mAP & ACC & F1 & A$\rightarrow$I (MRR) & I$\rightarrow$A (MRR) \\
\midrule
Supervise & 0.5300 & 0.6289 & 0.3212 & 0.4512 & 0.3827 \\
OpenL3 \cite{arandjelovic2017look} & 0.723 & 0.7681 & 0.4053 & 0.3405 & 0.651 \cite{9415085} & 0.1176 & 0.0171 & 0.0160 \\
YamNet & 0.8505 & 0.7832 &  \textbf{0.5039} & 0.4541 & 0.5667 & 0.3420 \\
Wav2CLIP \cite{wu2022wav2clip} & 0.8595 & 0.8101 & 0.4308 & 0.4663 &  0.6302 & 0.3955 & 0.0566 & 0.0678 \\
WavBriVL \cite{fang2023exploring} & 0.9117 & 0.8832 & - & 0.4741 &  - & 0.3720 & 0.0611 & 0.0608 \\
UniBriVL & 0.9307 & 0.8722 & - & 0.4885 &  - & 0.4111 & \textbf{0.0641} & 0.0612 \\
BGTAI & \textbf{0.9455} & \textbf{0.8911} & 0.4748 & \textbf{0.5130} & \textbf{0.6932} & \textbf{0.4351} & 0.0622 & \textbf{0.0745} \\
\midrule
BGTAI (ZS) & 0.5015 & 0.4896 & 0.0365 & 0.1214 & 0.2089 \\
\midrule
SOTA & 0.959 \cite{guzhov2021audioclip} & 0.8949 \cite{guzhov2021audioclip} &  0.5671 \cite{gong2021psla} & 0.544 \cite{Kazakos2021SlowFastAuditory} & 0.687 \cite{Fedorishin2021} \\
\bottomrule
\end{tabular}
\caption{The downstream classification and retrieval tasks' results are compared with supervised training from scratch, OpenL3, YamNet, Wav2CLIP, WavBriVL, UniBriVL and current state-of-the-art (SOTA) models. Zero-shot classification uses the BGTAI model.}
\label{tab:all_tasks}
\vspace{-1.0em}
\end{table*}

\begin{table}[htb]
\centering
\begin{tabular}{@{}cc@{\hskip 0.075in}c@{\hskip 0.075in}c@{\hskip 0.075in}c@{\hskip 0.075in}c@{\hskip 0.075in}c@{\hskip 0.075in}c@{\hskip 0.075in}c@{}}\toprule
Model & B1 & B4 & M & RL & Cr & S & Sr \\
\midrule
Baseline\footnotemark[8] \cite{Drossos_2020_icassp} & 0.389 & 0.015 & 0.084 & 0.262 & 0.074 & 0.033 & 0.054 \\
Wav2CLIP & 0.393 & 0.054 & 0.104 & 0.271 & 0.100 & 0.045 & 0.073 \\
WavBriVL & 0.412 & 0.92 & 0.108 & 0.268 & 0.126 & - & - \\
UniBriVL & 0.434 & 0.107 & 0.115 & 0.300 & 0.11 & 0.050 & 0.080 \\
BGTAI & \textbf{0.476} & \textbf{0.185} & \textbf{0.126} & \textbf{0.328} & \textbf{0.121} & \textbf{0.054} & \textbf{0.088} \\
\bottomrule
\end{tabular}
\caption{The audio captioning results are compared to the baseline, omitting Bleu2/3 and including Bleu1/4 (B1/4), METEOR (M), ROUGEL (RL), CIDEr (Cr), SPICE (S), and SPIDEr (Sr) metrics due to space limitations.}
\label{tab:audio_captioning}
\vspace{-12pt}
\end{table}

\subsection{Evaluation for Performance}

\subsubsection{Baseline Comparison}
\vspace{-0.1cm}
Table \ref{tab:all_tasks} compares the results of different models on classification and retrieval tasks. Models like OpenL3, YamNet, Wav2CLIP, WavBriVL, and UniBriVL achieve varying levels of accuracy on different datasets. The BGTAI model performs well in most tasks, with the highest accuracy on ESC-50 and UrbanSound8K datasets. In retrieval tasks, DESED (AR) evaluates based on the F1 score, where BGTAI achieves the highest score. For audio-visual cross-modal retrieval, UniBriVL achieves the highest MRR for A→I and BGTAI for I→A. 
The results presented in Table \ref{tab:all_tasks} demonstrate the effectiveness of different models on downstream classification and retrieval tasks. The BGTAI model consistently achieves competitive performance, outperforming other models on certain tasks. 

Table \ref{tab:audio_captioning} compares the results of different models for audio captioning using multiple evaluation metrics. The proposed models outperform the baseline. Wav2CLIP shows improved Bleu scores and higher METEOR and ROUGEL scores. WavBriVL performed well in most experiments and significantly improved METEOR. The UniBriVL model also performs well, particularly in terms of METEOR and other metrics. Finally, the BGTAI model outperforms all others, achieving the highest scores across all metrics. Most of the data came from the original paper, and we tried to keep the Settings consistent with the original paper in all tests. It's worth noting that some previous models have been fine-tuned on specific tasks that we haven't, which may result in our advantage not being as obvious as it should be. Overall, the proposed method and models demonstrate superior performance and potential in audio captioning tasks.

\begin{figure}[ht]
\vspace{-6pt}
\includegraphics[width=\linewidth]{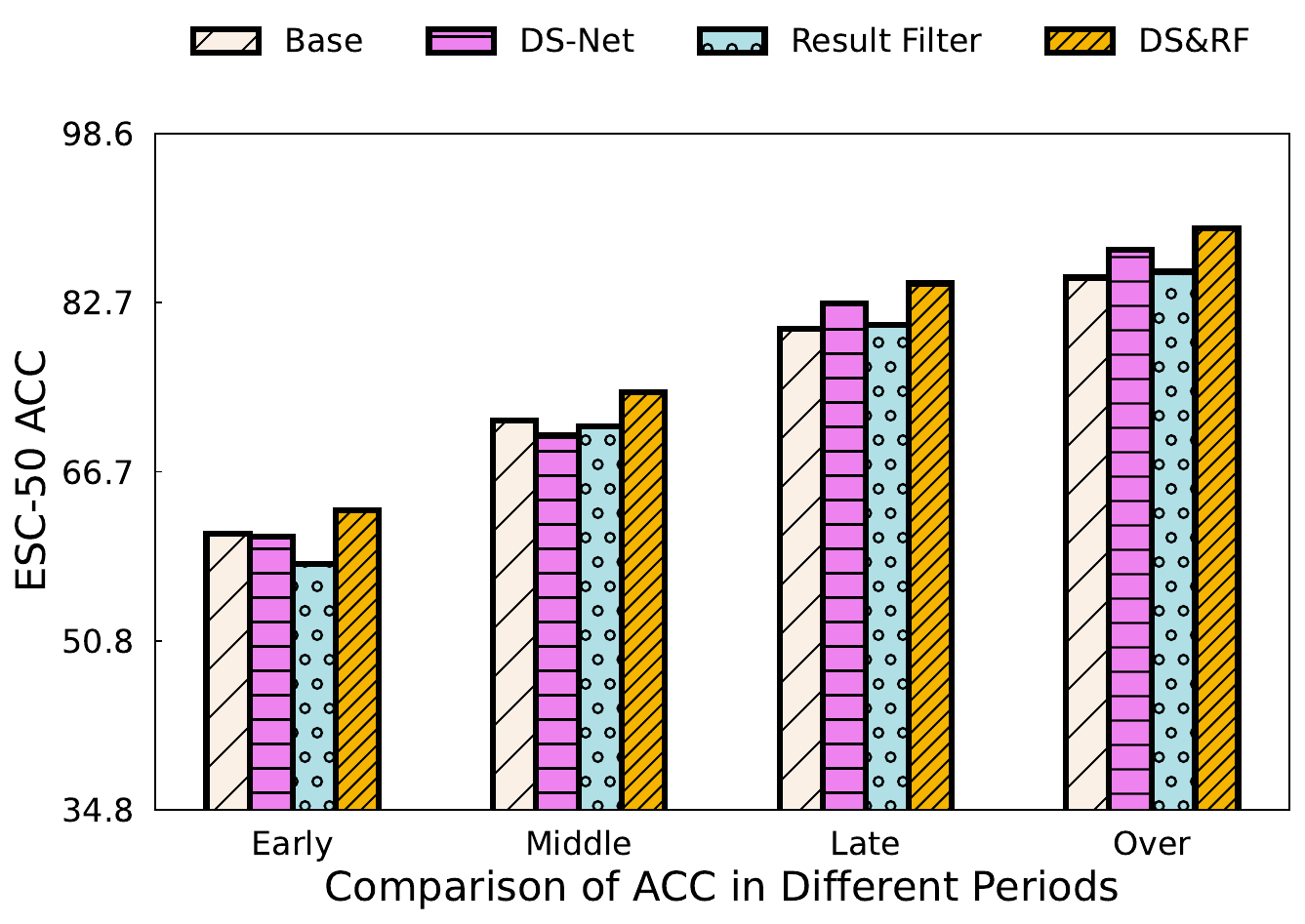}
\vspace{-16pt}
\caption{\textbf{Training efficiency evaluation.} We tested the performance of models under different settings on the ESC-50 dataset. We sampled the performance of different periods, determined by epoch.}
\label{fig:train}
\vspace{-16pt}
\end{figure}

\subsubsection{Training Efficiency Study}

According to the Fig \ref{fig:train}, the Base model exhibits the lowest ACC values across all periods, indicating its relatively lower efficiency in training. On the other hand, the set to "DS-Net" and "Result Filter" models show similar ACC values, indicating their comparable training efficiency.
Interestingly, the set to "DS\&RF" model demonstrates consistently higher ACC values across all periods, suggesting its superior training efficiency compared to the other models.
In summary, the comparison of ACC values in different periods clearly indicates that the set to "DS\&RF" model outperforms the other model configs in terms of training efficiency. 

We found the reason based on the results of the observation experiment, as shown in the bar chart, the "Result Filter" performance is not particularly impressive due to its comparatively long training time requirement. On the other hand, "DS-Net" is capable of eliminating poor data, resulting in a shorter training time requirement and consequently better performance within the same period. When the two are combined, they produce a better result than if set alone. Based on our speculation, theoretically, they perform better because they complement each other's shortcomings when combined: when used at the same time, the training time remains the same, while the overall efficiency performance becomes better than the individual.

\begin{figure*}[ht]
\vspace{-6pt}
\includegraphics[width=\linewidth]{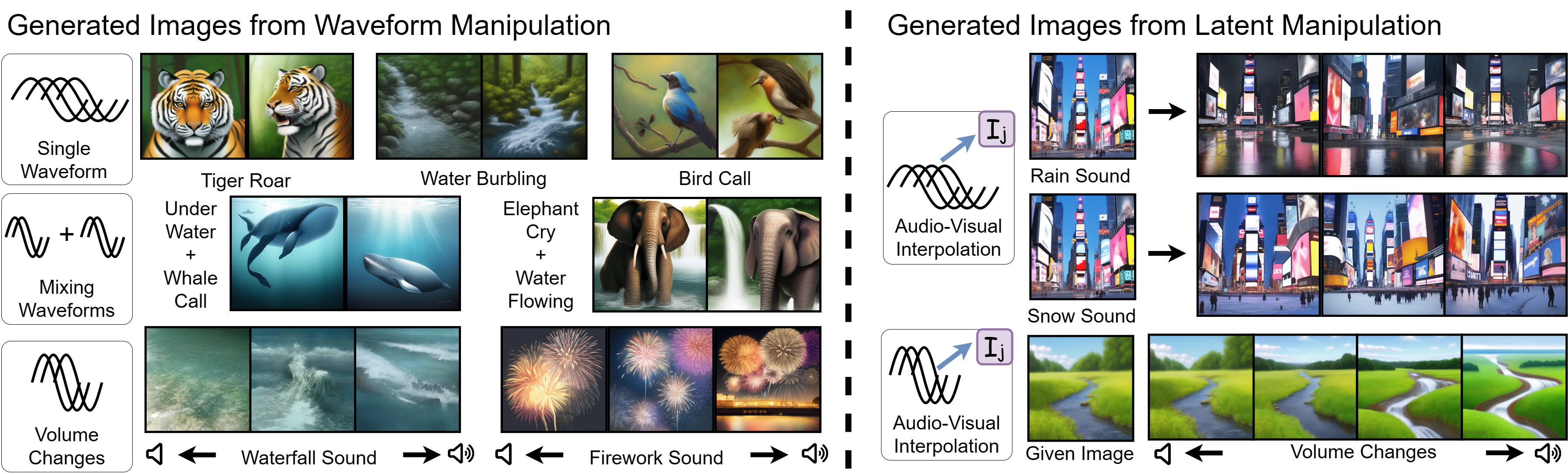}
\caption{{\bf Sound-to-image generation.} Our model introduces a groundbreaking approach for synthesizing natural scene images from sound. It is trained solely on paired audio-visual data, eliminating the need for labels or language supervision. Importantly, our model offers remarkable controllability by manipulating input waveforms (left) and controlling the generated images in the latent space of diffusion models \cite{Rombach_2022_CVPR} (right), the $I_{j}$ indicates image embedding. This innovative methodology provides enhanced flexibility and control over the model's outputs. 
}
\label{fig:S2I}
\vspace{-16pt}
\end{figure*}

\begin{table}[t]
\footnotesize
\centering
    \resizebox{1\linewidth}{!}{
    \begin{tabular}{l@{\quad}l@{\quad}c@{\quad}c@{\quad}c@{\quad}c@{\quad}c@{\quad}c}
    \toprule
    \multirow{2}{*}{}&\multirow{2}{*}{Loss}&\multirow{2}{*}{$D$}
    &\multirow{2}{*}{Duration}
    &\multicolumn{4}{c}{VGGSound (50 classes)}\\
   
    \cmidrule{5-8}
     & &&& R@1 & R@5 & FID ($\downarrow$) & IS ($\uparrow$)\\
    \cmidrule{1-8}
    (A) &$L_2$ & \checkmark&10 sec.& 14.57 & 37.35 & 19.24 & 7.98\\
    (B) & $L_{SP-Loss}$& \checkmark&10 sec.& 30.05 & 62.74 & 25.70 & 12.27\\
    (C) & $L_{total}$& & 10 sec.&36.10 & 70.89 & 20.52 & 17.00\\
    \cmidrule{1-8}
    (D) &\multirow{2}{*}{$L_{total}$}&\multirow{2}{*}{\checkmark}&1 sec. &34.78 & 73.62 & 19.37 & 19.25\\
    (E) &&&5 sec. &39.00 & 79.51 & 21.56 & 20.16\\
    \cmidrule{1-8}
    (F) & $L_{total}$& \checkmark& 10 sec.& \textbf{37.89} & \textbf{81.22} & \textbf{16.88} & \textbf{20.96}\\
    \bottomrule
    \end{tabular}
    }
    \caption{\textbf{Ablation studies of our proposed method.} We compare the different configurations of our method by changing the loss functions, frame selection method, and duration of the audio. $D$ denotes the Data-Pair Selection Network. 
     }
     \vspace{-16pt}
\label{tab:ablation}
\end{table}

\vspace{-0.27cm}
\subsubsection{Ablation Study}
\vspace{-0.1cm}
Table \ref{tab:ablation} presents ablation studies of our proposed method, evaluating different configurations by varying loss functions, frame selection, and audio duration. Configuration (B) improves the basic setup by incorporating the Self-Paced Learning loss, achieving higher performance. Removing the frame selection network and focusing on the total loss in configuration (C) further improves results. Surprisingly, reducing audio duration to 1 and 5 seconds in configurations (D) and (E) maintains high performance. Configuration (F) combines the advantages of previous setups and achieves the best performance, generating high-quality audio-visual embeddings.

\vspace{-0.25cm}
\subsubsection{Zero-Shot Comparison with Previous Works}
\vspace{-0.1cm}
Table \ref{tab:text_retrieval_audio} presents a zero-shot study on emergent audio retrieval and classification. The table compares methods based on their use of audio and text supervision or loss. 
In the first category, AudioCLIP achieves the highest Top-1 accuracy on the ESC dataset. AVFIC, in the second category, performs reasonably on Clotho and AudioCaps, but corresponding results for the ESC dataset are unavailable. In the third category, our proposed method, BGTAI, exhibits emergent capabilities and outperforms previous work on all datasets. In addition, we also provide performance comparisons with Imageblind, a relatively new multimodal large-scale model. Despite BGTAI having significantly fewer parameters, it achieves comparable performance to imageblind.
Overall, BGTAI demonstrates superior performance in emergent zero-shot audio retrieval and classification, surpassing existing methods with and without audio and text supervision.

\begin{table}[!t]
    \setlength{\tabcolsep}{3pt}
    \centering
    \resizebox{0.99\linewidth}{!}{
    \begin{tabular}{l c cc cc c}
    \toprule
    \multirow{2}{*}{Method}&\multirow{2}{*}{Emergent} & \multicolumn{2}{c}{\color{AudioDark} Clotho} &  \multicolumn{2}{c}{\color{AudioDark} AudioCaps} &  {\color{AudioDark} ESC}
    \\
    \cmidrule{3-7}
    && R@1 &  R@10  & R@1 & R@10 & Top-1
    \\
    \shline

    \multicolumn{3}{l}{\emph{Uses audio and text supervision}} \\
    AudioCLIP \cite{guzhov2021audioclip} & \xmark  & \_ & \_ & \_ & \_ & {\bf 68.6} \\ %
    \midrule
    \multicolumn{3}{l}{\emph{Uses audio and text loss}} \\
    AVFIC \cite{NEURIPS2021_76ba9f56} & \xmark &
    3.0 & 17.5 &  %
    8.7 &  37.7  %
    & \_ \\
    \midrule
    \multicolumn{3}{l}{\emph{No audio and text supervision}} \\
    \textbf{BGTAI (Ours)} & \cmark &
    \textbf{4.2} & \textbf{24.9} & \textbf{9.5} & \textbf{39.5} %
    & 62.3 %
    \\
    {\color{DarkGray}Imagebind} & {\color{DarkGray} \cmark} &
    {\color{DarkGray}6.0} & {\color{DarkGray}28.4} & {\color{DarkGray}9.3} & {\color{DarkGray}42.3} %
    & {\color{DarkGray}66.9} %
    \\
    \bottomrule
    \end{tabular}}
    \caption{\textbf{Zero-Shot Study.} Emergent zero-shot audio retrieval and classification, tests performed on Clotho \cite{Drossos_2020_icassp}, AudioCaps \cite{audiocaps} and ESC \cite{piczak2015dataset} datasets, models are all default Settings.}
    \label{tab:text_retrieval_audio}
    \vspace{-16pt}
\end{table}

\vspace{-0.27cm}
\subsection{Evaluation for Qualitative}
\vspace{-0.1cm}
As shown in Figure \ref{fig:S2I}, we conducted a qualitative evaluation to assess the capability of BGTAI in sound-guided image generation. It includes two major categories of demonstrations: direct manipulation of audio attributes such as volume and frequency, and audio embedding affecting image embedding. The evaluation demonstrates that our model successfully captures the intricate relationship between sound and visual content, translating auditory information into meaningful visual representations. Through these qualitative assessments, it is evident that our model achieves impressive performance in generating visual content from sound, offering exciting prospects for applications in creative industries and beyond. 
\vspace{-6pt}
\section{Conclusion}
\vspace{-4pt}
In this paper, we propose the BGTAI framework, which aims to bridge the gap between text, audio, images, and any other sequences in multimodal learning. The key component of our framework is the Langue2Gloss model, which represents text and audio inputs as gloss notations to better align them with images. We also introduce the DS-Net model, a Result Filter module, and a novel SP-Loss function to enhance the adaptability of gloss and improve training efficiency and stability. 
Our method outperforms previous multimodal models in various tasks, including audio retrieval, audio classification, audio captioning, and image retrieval. These findings suggest that the BGTAI framework has great potential for advancing multimodal learning and its application in real-world scenarios. 


\bibliographystyle{IEEEbib}
\bibliography{custom,others}

\begin{thebibliography}{10}

\bibitem{fei2022towards}
Nanyi Fei, Zhiwu Lu, Yizhao Gao, Guoxing Yang, Yuqi Huo, Jingyuan Wen, Haoyu
  Lu, Ruihua Song, Xin Gao, Tao Xiang, et~al.,
\newblock ``Towards artificial general intelligence via a multimodal foundation
  model,''
\newblock {\em Nature Communications}, vol. 13, no. 1, pp. 1--13, 2022.

\bibitem{wang2022image}
Wenhui Wang, Hangbo Bao, Li~Dong, Johan Bjorck, Zhiliang Peng, Qiang Liu, Kriti
  Aggarwal, Owais~Khan Mohammed, Saksham Singhal, Subhojit Som, et~al.,
\newblock ``Image as a foreign language: Beit pretraining for all vision and
  vision-language tasks,''
\newblock {\em arXiv preprint arXiv:2208.10442}, 2022.

\bibitem{girdhar2023imagebind}
Rohit Girdhar, Alaaeldin El-Nouby, Zhuang Liu, Mannat Singh, Kalyan~Vasudev
  Alwala, Armand Joulin, and Ishan Misra,
\newblock ``Imagebind: One embedding space to bind them all,'' 2023.

\bibitem{zhang2023metatransformer}
Yiyuan Zhang, Kaixiong Gong, Kaipeng Zhang, Hongsheng Li, Yu~Qiao, Wanli
  Ouyang, and Xiangyu Yue,
\newblock ``Meta-transformer: A unified framework for multimodal learning,''
  2023.

\bibitem{f2e24cc3-4208-3558-92fd-a3aa430647aa}
Samuel~J. Supalla, Jody~H. Cripps, and Andrew P.~J. Byrne,
\newblock ``Why american sign language gloss must matter,''
\newblock {\em American Annals of the Deaf}, vol. 161, no. 5, pp. 540--551,
  2017.

\bibitem{camgoz2020sign}
Necati~Cihan Camgoz, Oscar Koller, Simon Hadfield, and Richard Bowden,
\newblock ``Sign language transformers: Joint end-to-end sign language
  recognition and translation,''
\newblock in {\em IEEE Conference on Computer Vision and Pattern Recognition
  (CVPR)}, 2020.

\bibitem{Saunders_2022_CVPR}
Ben Saunders, Necati~Cihan Camgoz, and Richard Bowden,
\newblock ``Signing at scale: Learning to co-articulate signs for large-scale
  photo-realistic sign language production,''
\newblock in {\em Proceedings of the IEEE/CVF Conference on Computer Vision and
  Pattern Recognition (CVPR)}, June 2022, pp. 5141--5151.

\bibitem{fang2023unibrivl}
Sen Fang, Bowen Gao, Yangjian Wu, Jingwen Cai, and Teik~Toe Teoh,
\newblock ``Unibrivl: Robust universal representation and generation of audio
  driven diffusion models,'' 2023.

\bibitem{zhang2023speechlm}
Ziqiang Zhang, Sanyuan Chen, Long Zhou, Yu~Wu, Shuo Ren, Shujie Liu, Zhuoyuan
  Yao, Xun Gong, Lirong Dai, Jinyu Li, and Furu Wei,
\newblock ``Speechlm: Enhanced speech pre-training with unpaired textual
  data,'' 2023.

\bibitem{Sharma2022}
Roshan Sharma, Shruti Palaskar, Alan~W Black, and Florian Metze,
\newblock ``End-to-end speech summarization using restricted self-attention,''
\newblock in {\em ICASSP 2022 - 2022 IEEE International Conference on
  Acoustics, Speech and Signal Processing (ICASSP)}, 2022, pp. 8072--8076.

\bibitem{berndt1994dtw}
Donald~J Berndt and James Clifford,
\newblock ``{Using Dynamic Time Warping to Find Patterns in Time Series},''
\newblock in {\em AAA1-94 Workshop on Knowledge Discovery in Databases}, 1994.

\bibitem{piczak2015dataset}
Karol~J. Piczak,
\newblock ``{ESC}: {Dataset} for {Environmental Sound Classification},''
\newblock in {\em ACM Multimedia}. 2015, p. 1015, {ACM Press}.

\bibitem{Salamon:UrbanSound:ACMMM:14}
J.~Salamon, C.~Jacoby, and J.~P. Bello,
\newblock ``A dataset and taxonomy for urban sound research,''
\newblock in {\em ACM Multimedia}, Orlando, FL, USA, Nov. 2014, pp. 1041--1044.

\bibitem{fonseca2020fsd50k}
Eduardo Fonseca, Xavier Favory, Jordi Pons, Frederic Font, and Xavier Serra,
\newblock ``Fsd50k: an open dataset of human-labeled sound events,''
\newblock {\em arXiv preprint arXiv:2010.00475}, 2020.

\bibitem{chen2020vggsound}
Honglie Chen, Weidi Xie, Andrea Vedaldi, and Andrew Zisserman,
\newblock ``Vggsound: A large-scale audio-visual dataset,''
\newblock in {\em ICASSP}. IEEE, 2020, pp. 721--725.

\bibitem{9415085}
Shanshan Wang, Annamaria Mesaros, Toni Heittola, and Tuomas Virtanen,
\newblock ``A curated dataset of urban scenes for audio-visual scene
  analysis,''
\newblock in {\em ICASSP}, 2021, pp. 626--630.

\bibitem{Turpault2019_DCASE}
Nicolas Turpault, Romain Serizel, Ankit Parag~Shah, and Justin Salamon,
\newblock ``{Sound event detection in domestic environments with weakly labeled
  data and soundscape synthesis},''
\newblock in {\em {DCASE}}, New York City, United States, October 2019.

\bibitem{Drossos_2020_icassp}
Konstantinos Drossos, Samuel Lipping, and Tuomas Virtanen,
\newblock ``Clotho: {An} audio captioning dataset,''
\newblock in {\em ICASSP}, May 2020.

\bibitem{chen2015microsoft}
Xinlei Chen, Hao Fang, Tsung-Yi Lin, Ramakrishna Vedantam, Saurabh Gupta, Piotr
  Dollar, and C.~Lawrence Zitnick,
\newblock ``Microsoft coco captions: Data collection and evaluation server,''
  2015.

\bibitem{arandjelovic2017look}
Relja Arandjelovic and Andrew Zisserman,
\newblock ``Look, listen and learn,''
\newblock in {\em ICCV}. IEEE, 2017, pp. 609--617.

\bibitem{wu2022wav2clip}
Ho-Hsiang Wu, Prem Seetharaman, Kundan Kumar, and Juan~Pablo Bello,
\newblock ``Wav2clip: Learning robust audio representations from clip,''
\newblock in {\em ICASSP 2022-2022 IEEE International Conference on Acoustics,
  Speech and Signal Processing (ICASSP)}. IEEE, 2022, pp. 4563--4567.

\bibitem{fang2023exploring}
Sen Fang, Yangjian Wu, Bowen Gao, Jingwen Cai, and Teik~Toe Teoh,
\newblock ``Exploring efficient-tuned learning audio representation method from
  brivl,'' 2023.

\bibitem{guzhov2021audioclip}
Andrey Guzhov, Federico Raue, J{\"o}rn Hees, and Andreas Dengel,
\newblock ``Audioclip: Extending clip to image, text and audio,''
\newblock {\em arXiv preprint arXiv:2106.13043}, 2021.

\bibitem{gong2021psla}
Yuan Gong, Yu-An Chung, and James Glass,
\newblock ``Psla: Improving audio tagging with pretraining, sampling, labeling,
  and aggregation,''
\newblock {\em arXiv preprint arXiv:2102.01243}, 2021.

\bibitem{Kazakos2021SlowFastAuditory}
Evangelos Kazakos, Arsha Nagrani, Andrew Zisserman, and Dima Damen,
\newblock ``Slow-fast auditory streams for audio recognition,''
\newblock in {\em ICASSP}, 2021, pp. 855--859.

\bibitem{Fedorishin2021}
Dennis Fedorishin, Nishant Sankaran, Deen Mohan, Justas Birgiolas, Philip
  Schneider, Srirangaraj Setlur, and Venu Govindaraju,
\newblock ``Investigating waveform and spectrogram feature fusion for audio
  classification,''
\newblock Tech. {R}ep., DCASE2021 Challenge, June 2021.

\bibitem{Rombach_2022_CVPR}
Robin Rombach, Andreas Blattmann, Dominik Lorenz, Patrick Esser, and Bj\"orn
  Ommer,
\newblock ``High-resolution image synthesis with latent diffusion models,''
\newblock in {\em Proceedings of the IEEE/CVF Conference on Computer Vision and
  Pattern Recognition (CVPR)}, June 2022, pp. 10684--10695.

\bibitem{NEURIPS2021_76ba9f56}
Arsha Nagrani, Shan Yang, Anurag Arnab, Aren Jansen, Cordelia Schmid, and Chen
  Sun,
\newblock ``Attention bottlenecks for multimodal fusion,''
\newblock in {\em Advances in Neural Information Processing Systems},
  M.~Ranzato, A.~Beygelzimer, Y.~Dauphin, P.S. Liang, and J.~Wortman Vaughan,
  Eds. 2021, vol.~34, pp. 14200--14213, Curran Associates, Inc.

\bibitem{audiocaps}
Chris~Dongjoo Kim, Byeongchang Kim, Hyunmin Lee, and Gunhee Kim,
\newblock ``Audiocaps: Generating captions for audios in the wild,''
\newblock in {\em NAACL-HLT}, 2019.

\end{thebibliography}

\end{document}